# ConvNet Architecture Search for Spatiotemporal Feature Learning

Du Tran[1], Jamie Ray[1], Zheng Shou[2], Shih-Fu Chang[2], Manohar Paluri[1]
[1]Applied Machine Learning, Facebook, [2] Columbia University
{trandu,jamieray,mano}@fb.com  {zs2262,sc250}@columbia.edu

## Abstract

*Learning image representations with ConvNets by pre-training on ImageNet has proven useful across many visual understanding tasks including object detection, semantic segmentation, and image captioning. Although any image representation can be applied to video frames, a dedicated spatiotemporal representation is still vital in order to incorporate motion patterns that cannot be captured by appearance based models alone. This paper presents an empirical ConvNet architecture search for spatiotemporal feature learning, culminating in a deep 3-dimensional (3D) Residual ConvNet. Our proposed architecture outperforms C3D by a good margin on Sports-1M, UCF101, HMDB51, THUMOS14, and ASLAN while being 2 times faster at inference time, 2 times smaller in model size, and having a more compact representation.*

## 1. Introduction

Improving the design of ConvNet architectures has spurred significant progress in image understanding, with AlexNet [17] succeeded by VGG [34] and GoogleNet [38], then ResNet [8]. While video understanding is another fundamental problem in computer vision, the progress in architectures for video classification [14, 24, 33] and representation learning [41] is slower. There are three sources of friction impeding development of strong architectures for video. First, compared with image models, **video ConvNets have higher computation and memory cost**. For example, according to [41], it takes 3 to 4 days to train a 3D ConvNet on UCF101 and about two months on Sports-1M, which makes extensive architecture search difficult even on UCF101. Second, **there is not a standard benchmark to use for video architecture search**. In the static image setting, ConvNets can be trained on ImageNet [28] within a reasonable amount of time, and architectures that perform well on Imagenet have been shown to generalize to other tasks like object detection and segmentation. In the video domain, Sports-1M is shown to be helpful for generic feature learning [41], but is still too large to conduct architecture search. In contrast, while UCF101 has a similar number of frames to ImageNet, they are highly correlated and the setting is tightly controlled. As a result, models trained on this benchmark overfit easily; experiments in [41, 14] showed that ConvNets trained from scratch can obtain $41-44\%$ while finetuning from Sports1M can improve accuracy to $82\%$ [41] on UCF101. Third, **designing a video classification model is nontrivial**; there are many choices to which resulting performance is sensitive. These include how to sample and pre-process the input, what type of convolutions, how many layers to use, and how to model the temporal dimension (e.g. joint spatiotemporal modeling or decouple spatial and temporal dimensions). Thus, while it's clear that progress in the image domain should be incorporated in video modeling, a naive transfer of image models to video classification (e.g. simply applying a 2D Resnet to video frames) is suboptimal.

In this paper, we address these issues by conducting a carefully-designed architecture search on a small benchmark (UCF101). One might argue that the generalization of these findings is limited by the bias of the dataset - essentially, overfitting the search to UCF101. We tackle this difficulty via two efforts. First, we constrain networks to have similar capacity (number of parameters) - they will still overfit, but the improvements in accuracy can be more confidently attributed to a single change in architecture, not capacity. Second, the observations on this small-dataset architecture search lead us to an efficient deep 3D Residual ConvNet architecture (that we term Res3D), which we show to be effective when trained on a much larger dataset (Sports-1M), producing strong results on different video benchmarks. In summary, this paper makes the following contributions:

- We conduct a ConvNet architecture search across multiple dimensions by training on the UCF101 action recognition task, and present empirical observations of sensitivity to each dimension (section 3).

- We propose (to our best knowledge) the first deep 3D Residual network and train it on a large-scale video benchmark for spatiotemporal feature learning.

| Dataset | Sports1M acc.(%) | UCF101 acc.(%) | HMDB51 acc.(%) | THU14 mAP(%) | ASLAN acc.(%) |
|---|---|---|---|---|---|
| C3D | 61.1 | 82.3 | 51.6 | 19.0 | 78.3 |
| **Res3D** | **65.6** | **85.8** | **54.9** | **22.5** | **78.8** |
| Δ | 4.5 | 3.5 | 3.3 | 3.5 | 0.5 |

Table 1. **Comparison between Res3D and C3D**. Res3D outperforms C3D across different benchmarks by a good margin. Res3D achieves the best performance on different benchmarks including Sports-1M (among methods without long-term modeling), UCF101 and HMDB51 (among methods using only RGB input), THUMOS14, and ASLAN. We note that the 4.5% improvement gap on Sports1M is significant as random chance on this benchmark is about 0.2%.

- Our spatiotemporal representation achieves state-of-the-art results on Sports-1M (when no long-term modeling is used), UCF101 and HMDB51 (when considering only RGB input), and competative performance on THUMOS14, and ASLAN.

- Our model is **2** times faster, **2** times smaller, and more compact than current deep video features.

## 2. Related Work

Video understanding is one of the core computer vision problems and has been studied for decades. Many research contributions in video understanding have focused on developing spatiotemporal features for videos. Some proposed video representations include spatiotemporal interest points (STIPs) [19], SIFT-3D [30], HOG3D [15], Cuboids [2], and ActionBank [29]. These representations are hand-designed and use different feature encoding schemes like feature histograms or pyramids. Among hand-crafted representations, improved Dense Trajectories (iDT) [42] is known as the current state-of-the-art hand-crafted feature with strong results on different video classification problems.

Since the deep learning breakthrough in computer vision [17] presented at the ImageNet Challenge 2012 [28], many ConvNet-based methods [21] were proposed for image recognition. Simonyan and Zisserman proposed to stack multiple small $3 \times 3$ kernels to approximate bigger kernels (e.g. $5 \times 5$ or $7 \times 7$) with more non-linear RELU units in between, and obtained good image classification performance [34] with an ConvNet architecture known as VGG. Various techniques have been developed to improve image classification including Batch Normalization [10], Parameterized-RELU [7], Spatial Pyramid Pooling [6]. Inspired by the idea of Network in Network [22], different GoogleNet (a.k.a. Inception) models were proposed with strong performance on ImageNet [37, 39, 38]. Recently, He *et al.* proposed deep residual networks (Resnets), which won multiple tracks in the ImageNet 2015 challenges [8]. By using residual connections, deeper ConvNets can be trained with much less overfitting.

Deep learning has also been applied to video understanding. 3D ConvNets were proposed for recognizing human actions in videos [11], and 3D convolutions were also used in Restricted Boltzmann Machines [40] and stacked ISA [20] to learn spatiotemporal features. Karpathy *et al.* [14] proposed different fusion methods for video classification. Simonyan and Zisserman [33] used two-stream networks to achieve high accuracy on action recognition. Feichtenhofer *et al.* enhanced these two-stream networks with Resnet architectures and additional connections between streams [4]. Some two-stream-network based approaches including Temporal Segment Networks [44], Action Transformations [45], and Convolutional Fusion [5] were proposed and achieved the best accuracy for human action recognition. Recently, Tran *et al.* proposed to train a deep 3D ConvNet architecture, called C3D, on a large-scale dataset for spatiotemporal feature learning [41]. The C3D features have strong performance on various tasks, including action recognition [41], action detection [32], video captioning [26], and hand gesture detection [23].

Our approach in this paper is mostly related to C3D [41] and Resnet [8]. Similar to C3D, we use 3D ConvNets to learn spatiotemporal features. However, while the work in [41] is limited to searching for the 3D convolution temporal kernel length, we consider many other dimensions of architecture design. Furthermore, our search is designed to compare different architectures while constraining the model capacity (number of parameters). Our work is also related to Resnet [8] in the way we constrain our search to Resnet architectures. However, we emphasize that the application of Resnet to video representation is challenging as we need to consider many nontrivial questions, which we answer empirically via carefully-designed experiments (section 3). Our newly proposed architecture (Res3D) outperforms C3D by a good margin across **5** different benchmarks while being **2**x faster in run-time and **2**x smaller in model size (section 4).

## 3. Architecture Search

In this section we present a large-scale search for ConvNet architectures amenable to spatiotemporal feature learning. We start with C3D [41] as it is commonly used as a deep representation for videos. We also constrain our search space to deep residual networks (Resnet [8]) owing to their good performance and simplicity. Due to the high computational cost of training deep networks, we conduct our architecture search on UCF101 [35] split 1. We also note that the high memory consumption of these networks, coupled with the need for a large minibatch to compute batch normalization statistics, prohibits exploring some parts of model space. The observations from these experiments are collectively adopted in designing our final proposed ConvNet architecture. We later train our final ar-chitecture on Sports-1M [14] and show the benefits of the learned spatiotemporal features across different video understanding tasks.

### 3.1. Remark on Generality

These are empirical findings on a small benchmark (UCF101). We attempt to limit the confounding effect of overfitting by constraining the capacity of each network in the search, so that differences in performance can be more confidently attributed to the design of a network, rather than its size. Nevertheless, one might ask whether these findings can generalize to other datasets (in particular, larger ones like Sports-1M). While it is prohibitive to replicate each experiment at larger scale, we will choose a few in following sections to show that the findings are consistent. We stress that while this protocol isn't ideal, it is practical and the resulting intuitions are valuable - nevertheless, we encourage the development of a benchmark more suited to architecture search.

### 3.2. 3D Residual Networks

**Notations**: For simplicity we omit the channels, and denote input, kernel, and output as 3D tensors of $L \times H \times W$, where $L$, $H$, and $W$ are temporal length, height, and width, respectively.

**Basic architectures**: Our basic 3D Resnet architectures are presented in Table 2. These networks use an $8 \times 112 \times 112$ input, the largest that can fit within GPU memory limits and maintain a large enough mini-batch. However, we skip every other frame, making this equivalent to the using C3D input and dropping the even frames. In summary, we modify 2D-Resnets by: changing the input from $224 \times 224$ to $8 \times 112 \times 112$; changing all convolutions from $d \times d$ to $3 \times d \times d$ with all downsampling convolution layers using stride $2 \times 2 \times 2$ except for conv1 with stride $1 \times 2 \times 2$; and removing the first max-pooling layer.

**Training and evaluation**: We train these networks on UCF101 train split 1 using SGD with mini-batch size of 20. Similar to C3D, video frames are scaled to $128 \times 171$ and randomly cropped to $112 \times 112$. Batch normalization is applied at all convolution layers. The learning rate is set to 0.01 and divided by 10 after every 20k of iterations. Training is done at 90K iterations (about 15 epochs). We also conduct 2D-Resnet baselines where we replace the input with a single frame cropped to $112 \times 112$ and all 3D operations (convolution and pooling) with 2D analogues. Table 3 presents the clip accuracy of different networks on UCF101 test split 1. Compared to the 2D reference models, the 3D-Resnets perform better, and the deeper networks (34 layers) show little gain over the 18-layer ones for 2D or 3D. We note that these findings are not conclusive as the networks have different number of parameters. In the following architecture search experiments, all compared models use a similar number ( 33M) of parameters.

| layer name | output size | 3D-Resnet18 | 3D-Resnet34 |
|---|---|---|---|
| conv1 | 8×56×56 | 3×7×7, 64, stride 1 × 2× 2 | |
| conv2_x | 8×56×56 | $\begin{bmatrix} 3\times3\times3, 64 \\ 3\times3\times3, 64 \end{bmatrix}\times2$ | $\begin{bmatrix} 3\times3\times3, 64 \\ 3\times3\times3, 64 \end{bmatrix}\times3$ |
| conv3_x | 4×28×28 | $\begin{bmatrix} 3\times3\times3, 128 \\ 3\times3\times3, 128 \end{bmatrix}\times2$ | $\begin{bmatrix} 3\times3\times3, 128 \\ 3\times3\times3, 128 \end{bmatrix}\times4$ |
| conv4_x | 2×14×14 | $\begin{bmatrix} 3\times3\times3, 256 \\ 3\times3\times3, 256 \end{bmatrix}\times2$ | $\begin{bmatrix} 3\times3\times3, 256 \\ 3\times3\times3, 256 \end{bmatrix}\times6$ |
| conv5_x | 1×7×7 | $\begin{bmatrix} 3\times3\times3, 512 \\ 3\times3\times3, 512 \end{bmatrix}\times2$ | $\begin{bmatrix} 3\times3\times3, 512 \\ 3\times3\times3, 512 \end{bmatrix}\times3$ |
| | 1×1×1 | average pool, 101-d fc, softmax | |

Table 2. **Basic 3D-Resnet architectures**. Building blocks are shown in brackets, with the numbers of blocks stacked. Downsampling is performed by conv3_1, conv4_1, and conv5_1 with a striding of 2×2×2.

| Net | 2D-Res18 | 2D-Res34 | 3D-Res18 | 3D-Res34 |
|---|---|---|---|---|
| # params ($\times 10^6$) | 11.2 | 21.5 | 33.2 | 63.5 |
| FLOPs ($\times 10^9$) | 1.6 | 3.5 | 19.3 | 36.7 |
| Accuracy (%) | 42.2 | 42.2 | 45.6 | 45.9 |

Table 3. **Accuracy on UCF101, varying Resnet**. 3D-Resnets achieve better accuracy compared to 2D ones, however the finding is not conclusive because the 3D-Resnets have many more parameters.

**Simplified networks**: We note that by reducing the input size, we can further reduce the complexity of the network and the memory consumption of network training, thus accelerating architecture search. With a smaller input of $4 \times 112 \times 112$, we have to adjust the stride of conv5_1 to $1 \times 2 \times 2$. This simplification reduces the complexity of 3D-Resnet18 from 19.3 billion floating point operations (FLOPs) to 10.3 billion FLOPs while maintaining accuracy on UCF101 (within the margin of random chance, $0.96\%$). From now on we denote this network architecture as SR18 (the Simplified 3D-Resnet18), and use it as a baseline to modify for our following architecture search experiments.

**Observation 1**. Using 4 frames of input and a depth-18 network (SR18) achieves good baseline performance and fast training on UCF101.

### 3.3. What are good frame sampling rates?

We use SR18 and vary the temporal stride of the input frames in the following set $\{1, 2, 4, 8, 16, 32\}$. At the lowest extreme the input is 4 consecutive frames which is roughly a 1/8-second-long clip. On the other hand, using stride 32 the input clip is coarsely sampled from a 128-frame long clip ($\sim$ 4.5 to 5 seconds). Table 4 presents the accuracy of SR18 trained on inputs with different sampling rates.

| Input Stride | 1 | 2 | 4 | 8 | 16 | 32 |
|---|---|---|---|---|---|---|
| Accuracy (%) | 43.8 | 46.3 | 46.1 | 38.8 | 38.1 | 36.9 |

Table 4. **Accuracy on UCF101, varying sampling rates**. Temporal strides between 2 and 4 are reasonable, while 1 is a bit less accurate, suggesting these clips are too short and lack context. On the other hand, strides beyond 4 are significantly worse, maybe because they lack coherence and make it hard for 3D kernels to learn temporal information.

| Input resolution (crop size) | 64 (56) | 128 (112) | 256 (224) |
|---|---|---|---|
| Accuracy (%) | 37.6 | 46.1 | 44.0 |
| FLOPs ($\times 10^9$) | 2.6 | 10.3 | 42.9 |
| # params ($\times 10^6$) | 33.2 | 33.2 | 33.3 |

Table 5. **Accuracy on UCF101, varying input resolution**. With input resolution of 64, SR18 accuracy drops 9.5%, and 128 gives higher accuracy and less computation than 256. It's possible that at high resolution the larger `conv1` kernels are harder to learn.

**Observation 2**. For video classification, sampling one frame out of every 2-4 (for videos within 25-30fps), and using clip lengths between 0.25s and 0.75s yields good accuracy.

### 3.4. What are good input resolutions?

In [41], Tran *et al*. conducted an experiment to explore the input resolutions. However, their networks have different numbers of parameters, thus different levels of overfitting can be expected. We conduct a similar experiment to determine a good input resolution for video classification, but again constrain our networks to use a similar number of parameters. We experiment with 3 different input resolutions of $224 \times 224$, $112 \times 112$, and $56 \times 56$ with re-scaled frame size $256 \times 342$, $128 \times 171$, and $64 \times 86$, respectively. We adjust the kernel size of the `conv1` layers of these networks so that they have similar receptive fields, using $3 \times 11 \times 11$, $3 \times 7 \times 7$, $3 \times 5 \times 5$ with stride of $1 \times 4 \times 4$, $1 \times 2 \times 2$, $1 \times 1 \times 1$ for input resolution 224, 112, and 56, respectively. The rest of these three networks remains the same, thus the only difference in parameters comes from the conv1 layer which is small compared with the total number of parameters. Table 15 reports the accuracy of these different input resolutions.

**Observation 3**. An input resolution of 128 (crop 112) is ideal both for computational complexity and accuracy of video classification given the GPU memory constraint.

### 3.5. What type of convolutions?

There are many conjectures about the type of convolutions used for video classification. One can choose to use 2D ConvNets as in two stream networks [33] or fully 3D ConvNets as in C3D [41], or even a ConvNet with mixtures of 2D and 3D operations [14]. In this section, we compare a mixed 3D-2D ConvNet and mixed 2D-1D ConvNet with a full-3D ConvNet (SR18) to address these conjectures.

**Mixed 3D-2D ConvNets**. One hypothesis is that we only need motion modeling at some low levels (early layers) while at higher levels of semantic abstraction (later layers), motion or temporal modeling is not necessary. Thus some plausible architectures may start with 3D convolutions and switch to using 2D convolutions at the top layers. As SR18 has 5 groups of convolutions, our first variation is to replace all 3D convolutions in group 5 by 2D ones. We denote this variant as MC5 (mixed convolutions). Similarly, we replace group 4 and 5 with 2D convolutions, and name the variation is MC4 (meaning from group 4 and deeper layers all convolutions are 2D). Following this pattern, we also create MC3, MC2, and MC1 variations. We note that MC1 is equivalent to a 2D ConvNet applied on clip inputs, which differs slightly from the architecture presented in Table 2, which is a 2D-Resnet18 applied on frame inputs. In order to constrain the model capacity, we fix the number of filters in each group of conv1_x to conv5_x to be $k, k, 2k, 4k, 8k$, respectively, and call $k$ the network width. When 3D convolution kernels ($3 \times d \times d$) are replaced by 2D convolutions ($1 \times d \times d$), there is a reduction in the number of parameters; we can then adjust $k$ as needed to keep the capacity comparable to SR18.

**2.5D ConvNets**. Another hypothesis is that 3D convolutions aren't needed at all, as 3D convolutions can be approximated by a 2D convolution followed by a 1D convolution - decomposing spatial and temporal modeling into separate steps. We thus design a network architecture that we call a 2.5D ConvNet, where we replace each $3 \times d \times d$ 3D convolution having $n$ input and $m$ output channels by a 2.5D block consisting of a $1 \times d \times d$ 2D convolution and a $3 \times 1 \times 1$ 1D convolution layer having $i$ internal channel connections such that the number of parameters are comparable (Figure 1). If the 3D convolution has spatial and temporal striding (e.g. downsampling), the striding is also decomposed according to its corresponding spatial or temporal convolutions. We choose $i = \lceil \frac{3mnd^2}{nd^2+3m} \rceil$ so that the number of parameters in the 2.5D block is approximately equal to that of the 3D convolution. This is similar to the bottleneck block in [8] which factorized two 2D convolutions into two 1D and one 2D convolutions. Table 16 reports the accuracy of these convolution variations.

**Observation 4**. Using 3D convolutions across all layers seems to improve video classification performance.

### 3.6. Is your network deep enough?

Here, we use SR18 as a reference model and vary the network depth with the same constraint on number of parameters. We again design the network width $k$ such that all convolution layers in group 1 and 2 (`conv1` and `conv2_x`) have $k$ output filters, `conv_3x`, `conv_4x`, and `conv_5x` have $2k$, $4k$, and $8k$ output filters, respectively. By changing the number of blocks in each group we obtain networks of different depth, and again adjust $k$ to match the number of parameters.

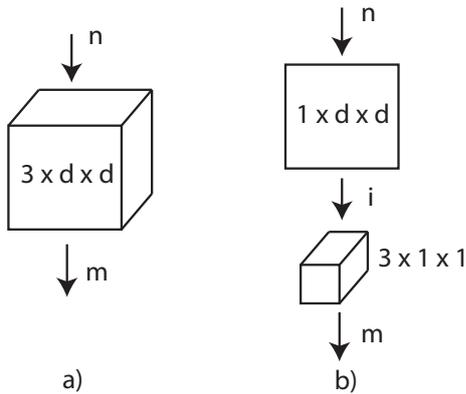

Figure 1. **2.5D Convolution Block vs. 3D Convolution**. a) A 3D Convolution layer with $n$ input channels and $m$ output channels. b) a mixed 2D and 1D convolution block with the same $n$ input and $m$ output channels. $d$ is the spatial size of the kernels. $i$ is the output channels for the 2D convolution layer and also the input channels for 1D convolution layer, and is chosen to make the number of parameters in both blocks equal.

| Net | MC1 | MC2 | MC3 | MC4 | MC5 | 2.5D | 3D |
|---|---|---|---|---|---|---|---|
| k | 109 | 109 | 108 | 103 | 90 | - | 64 |
| # params ($\times 10^6$) | 32.5 | 32.5 | 32.8 | 32.4 | 33.0 | 33.2 | 33.2 |
| FLOPs ($\times 10^9$) | 10.2 | 9.7 | 20.8 | 22.6 | 18.7 | 10.3 | 10.3 |
| Accuracy (%) | 43.3 | 44.5 | 44.8 | 44.3 | 45.1 | 44.6 | 46.1 |

Table 6. **Accuracy on UCF101, varying convolution mixtures**. SR18 with fully 3D convolution layers performs the best. The accuracy gap is not very large, but still significant relative to random chance of 0.96%. $k$ is the network width, which we adjust so that all comparing networks have similar number of parameters.

| Net | Blocks | k | # params ($\times 10^6$) | FLOPs ($\times 10^9$) | Acc (%) |
|---|---|---|---|---|---|
| D10 | [1 1 1 1] | 98 | 33.7 | 10.8 | 45.3 |
| D16 | [2 2 2 1] | 85 | 33.6 | 16.7 | 46.8 |
| D18 | [2 2 2 2] | 64 | 33.2 | 10.3 | 46.1 |
| D26 | [2 3 4 3] | 50 | 33.8 | 8.4 | 46.3 |
| D34 | [3 4 6 3] | 46 | 32.8 | 10.0 | 45.8 |

Table 7. **Accuracy on UCF101, varying depth**. The second column denotes the numbers of blocks in each convolution group. $k$ is purposely selected to make the networks have similar number of parameters. D18 is our SR18 reference model. There is no clear winner among these networks, except that D34 and D10 are somewhat worse. This suggests that for the same number of parameters, the depth of 16-26 layers is good enough for video classification.

Table 17 reports the effects of model depth on accuracy. We note that He *et al.* [8] conducted a similar investigation of depth for ImageNet classification, but did not constrain the number of parameters of the comparing networks. Their experiments are on ImageNet which is large-scale, while UCF101 exhibits much more overfitting and is thus more sensitive to the model capacity. Although D16, D18, and D26 provide a similar accuracy, D18 has a lower complexity than D16, and consumes less memory than D26.

**Observation 5**. A network depth of 18 layers gives a good trade-off between accuracy, computational complexity, and memory for video classification.

## 4. Spatiotemporal feature learning with 3D Resnets

We now apply the observations of section 3 to design networks for spatiotemporal feature learning on a large-scale dataset (Sports-1M). We then compare the learned features with the current C3D features on a variety of video tasks in section 5.

**Architecture**. We select the 3D-Resnet18 architecture (shown in Table 2) as suggested by the empirical observations in section 3. In contrast to SR18, we use an input of $8 \times 112 \times 112$ frames, because large-scale training can benefit from the additional information. All other observations are adopted: temporal stride of 2, input resolution of $112 \times 112$, full-3D convolutions with depth of 18. We denote this architecture as Res3D from now on.

**Training**. Similar to C3D, we train our Res3D on Sports-1M [14] to learn spatiotemporal features. Sports-1M is a large-scale dataset with about 1.1M videos of 487 fine-grained sports categories, and includes a public train and test split. We randomly extract five 2-second-long clips from each training video. Clips are resized to $128 \times 171$ resolution. Random spatiotemporal jittering is applied as data augmentation to randomly crop the input clips to $8 \times 112 \times 112$ (sampling stride 2). Training uses SGD on 2 GPUs with a mini-batch size of 40 examples (20 per GPU). The initial learning rate is 0.01 and is divided by 2 every 250k iterations, finishing at 3M iterations. We also train a 2D-Resnet18 baseline with the same procedure.

**Results on Sports1M**. Table 8 presents the classification results of our Res3D compared with current methods. For top-1 clip accuracy we use a single center-cropped clip and a single model. For video top-1 and top-5 accuracy, we use 10 center-cropped clips and average their predictions to make a video-level prediction. Res3D achieves state-of-the-art performance when compared with single models that do not use long-term modeling. It outperforms the previous state of the art, C3D [41], by 2.7%, 4.5%, and 2.6% on top-1 clip, top-1 video, and top-5 video accuracy respectively. Compared with 2D ConvNets, Res3D improves 2% and 0.7% over AlexNet and GoogleNet on top-1 video accuracy. We note that these methods use 240 crops where Res3D uses only 10 crops. These improvements are significant compared to random chance (only 0.2%) on this large-scale benchmark. Compared to methods that use long-term modeling, e.g. LSTM or Convolution Pooling [24], our Res3D is 2.9% and 4.2% better than an LSTM trained on AlexNet fc features, and comparable with an LSTM trained on GoogleNet fc features. The only methods with higher accuracy use convolutional pooling for long-term modeling. We note that our method does not involve any long-term modeling because our main objective is to learn atomic spatiotemporal features. In fact, the orthogonal direction of long-term modeling with LSTM or convolution pooling can be applied on our Res3D model to further improve sports classification accuracy.

| Method | Clip@1 | Video@1 | Video@5 |
|---|---|---|---|
| single model, no long-term modeling | | | |
| DeepVideo [14] | 41.9 | 60.9 | 80.2 |
| C3D [41] | 46.1 | 61.1 | 85.2 |
| AlexNet [24] | N/A | 63.6 | 84.7 |
| GoogleNet [24] | N/A | 64.9 | 86.6 |
| 2D-Resnet* | 45.5 | 59.4 | 83.0 |
| **Res3D (ours)*** | **48.8** | **65.6** | **87.8** |
| with long-term modeling | | | |
| LSTM+AlexNet [24] | N/A | 62.7 | 83.6 |
| LSTM+GoogleNet [24] | N/A | 67.5 | 87.1 |
| Conv pooling+AlexNet [24] | N/A | 70.4 | 89.0 |
| Conv pooling+GoogleNet [24] | N/A | **71.7** | **90.4** |

Table 8. **Results on Sports-1M**. Upper table presents the sports classification accuracy of different methods when a single model is used. The lower table presents the results of the methods that use multiple crops and long-term modeling. Our Res3D achieves state-of-the-art accuracy when compared with methods that do not employ long-term modeling. The results of the other methods are quoted directly from the relevant papers. Note that random chance on this dataset is only 0.2%. * 2D-Resnet and Res3D use only $112 \times 112$ resolution while the others use $224 \times 224$.

Figure 2 visualizes the learned conv1 filters of both Res3D and the 2D-Resnet baseline. These two networks are trained on Sports-1M; the only difference is that 3D convolutions are replaced by 2D ones. We observe that: 1) All 3D filters change in the time dimension, meaning each encodes spatiotemporal information, (not just spatial); 2) For most of the 2D filters, we can find a 3D filter with a similar appearance pattern (mostly at the center of the kernel). This may indicate that 3D filters are able to cover the appearance information in 2D filters but can also capture useful motion information. This confirms our finding (consistent with [41]) that 3D convolutions can capture appearance and motion simultaneously, and are thus well-suited for spatiotemporal feature learning.

**Model size and complexity**. Res3D is about 2 times smaller and also 2 times faster than C3D. Res3D has 33.2 million parameters and 19.3 billion FLOPs while C3D has 72.9 million parameters and 38.5 billion FLOPs.

| Model | Classifier | HMDB51 | UCF101 |
|---|---|---|---|
| 2D-Resnet | linear SVM | 47.2 | 79.2 |
| C3D [41] | linear SVM | 51.6 | 82.2 |
| C3D [41] | fine-tuned | 50.3 | 82.3 |
| **Res3D (ours)** | linear SVM | 51.3 | 83.1 |
| **Res3D (ours)** | fine-tuned | **54.9** | **85.8** |

Table 9. **Action recognition results on HMDB51 and UCF101**. Res3D outperforms C3D by 3.5% on UCF101 and 3.3% on HMDB51.

## 5. Res3D as Spatiotemporal Features

In this section, we evaluate our Res3D model pre-trained on Sports-1M as a spatiotemporal feature extractor for video understanding tasks and compare with other representations.

### 5.1. Action recognition

**Datasets**. In this experiment, we use UCF101 [35] and HMDB51 [18], which are popular public benchmarks for human action recognition. UCF101 has 13,320 videos and 101 different human actions. HMDB51 has about 7,000 videos and 51 human action categories. Both have 3 train/test splits, so we evaluate with 3-fold cross validation.

**Models**. There is one publicly available deep spatiotemporal feature: C3D [41]. We compare our Res3D with C3D and the 2D-Resnet baseline which was trained on Sports-1M frames. We use the fc6 activations of C3D as suggested by the authors [41]. For Res3D and 2D-Resnet, we tried both res5b and pool5. The res5b features are slightly better than pool5 for both 2D and 3D cases, but the gap is small; here we report the result using res5b. We represent a video by extracting clip features and average pooling them, then L2-normalizing the resulting vector. A linear SVM is used to classify the actions. We also fine-tune C3D and Res3D on both UCF101 and HMDB51 and find out that the performance gaps for fine-tuned models are bigger, suggesting that a stronger architecture benefits more from fine-tuning.

Table 9 shows the result on human action recognition on UCF101 and HMDB51 of Res3D compared with C3D [41] and the 2D-Resnet baseline. Our Res3D outperforms C3D by 3.5% and 3.3% on UCF101 and HMDB51, respectively. It achieves the best accuracy of methods that use only RGB input (see Table 10). Temporal Segment Networks (TSNs) also achieves very high accuracy of 85.7% on UCF101 when using only RGB. TSNs can even achive higher accuracy of 87.3% when using with two modalities (e.g. RGB and RGB Difference). We note that these numbers are evaluated on split 1 only, thus not directly compparable to our results. Although the direct comparion is not possible here, we can still conjecture that TSNs and Res3D architecture are in par when applied on the same RGB modality. On HMDB51, Res3D achieves the best performance among the methods using only RGB input. State of the art models [44, 5, 4] augment RGB input with expensive optical flow or iDT [45], but it is worth noting that the high computational cost of such augmentations prohibits their application at a large scale (e.g. Sports-1M). The very recent work, I3D [1] (concurrent with this work), achieves very good performance on UCF101 (98%) and HMDB51 (80%) with their two-stream I3D model using both RGB and optical flow inputs, and an Imagenet pre-trained model. When using only RGB, their results on UCF101 and HMDB51 are $84.5\%$ and $49.8\%$ which are $1.3\%$ and $5.1\%$ worse than ours. We note these results are not directly comparable as their number on only split 1 of UCF101 and HMDB51.

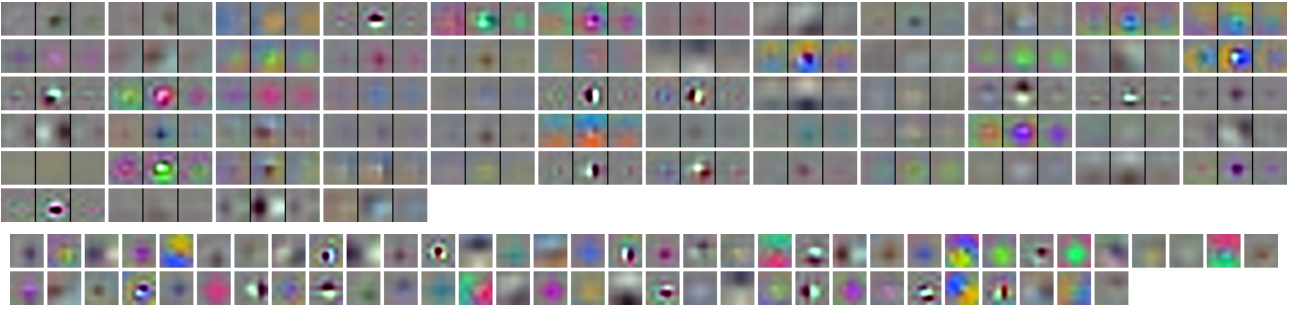

Figure 2. **Res3D vs. 2D-Resnet learned filters**. Visualization of the Res3D 64 filters in its `conv1` layer compared with those of the 2D-Resnet baseline. Both networks are trained on Sports1M. The upper images are Res3D filters ($3 \times 7 \times 7$) presented as a group of 3 images (upscaled by 4x). The lower images are 2D-Resnet filters ($7 \times 7$) upscaled by 4x. We see that both models capture appearance information (color, edges, texture), but the 3D filters also incorporate temporal dynamics, thus they are well-suited for spatiotemporal features. Best viewed in color. GIF annimations of 3D filters can be viewed here: http://goo.gl/uES8Ma.

| Method | UCF101 | HMDB51 |
|---|---|---|
| Slow Fusion [14] | 65.4 | - |
| Spatial Stream [33] | 73.0 | 40.5 |
| LSTM Composite Model [36] | 75.8 | 44.1 |
| Action Transformations [45] | 80.8 | 44.1 |
| C3D (1 net) | 82.3 | 51.6 |
| Conv Pooling [24] | 82.6 | - |
| Conv Fusion [5] | 82.6* | 47.1* |
| Spatial Stream-Resnet [4] | 82.3 | 43.4 |
| TSNs [44] (RGB) | **85.7*** | - |
| I3D [1] (RGB) | 84.5* | 49.8* |
| **Res3D (ours)** | **85.8** | **54.9** |

Table 10. Comparison with state-of-the-art methods on HMDB51 and UCF101 considering methods that use only RGB input. We note that the current state of the art is about 94%-98% and 69% on UCF101 and HMDB51 [1, 45, 44, 5, 4] when using optical flow and improved dense trajectories [42]. *Results are computed on only split 1.

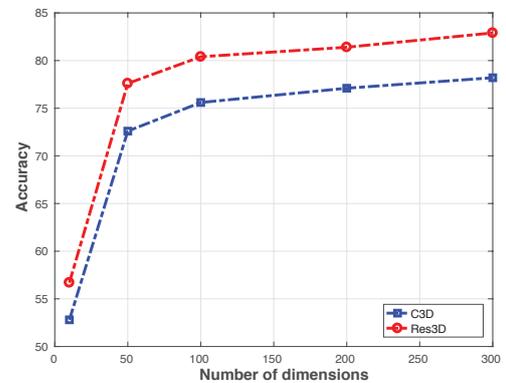

Figure 3. **Res3D is more compact than C3D**. Res3D outperforms C3D by 4 to 5% after PCA projection to low dimensions, showing that its feature is much more compact than C3D.

Figure 3 compares the accuracy (on UCF101) of Res3D with that of C3D at low dimensions. We use PCA to project the spatiotemporal features of C3D (fc6) and Res3D (res5b) to a low dimensionality and then classify them with a linear SVM. Res3D outperforms C3D by $3.9\%$ (56.7 vs. 52.8) at 10 dimensions, and improves about $5\%$ over C3D at 50, 100, 200, and 300 dimensions. This indicates that Res3D is much more compact compared with C3D. In fact, Res3D has accuracy of $82.9\%$ at only 300 dimensions which is already better than C3D using the full 4,096 dimensions.

### 5.2. Action similarity labelling

**Dataset**. The ASLAN dataset[16] has 3,631 videos from 432 action classes. The task is to predict if a given pair of videos contain the same or different action. We use the public 10-fold cross validation protocol.

**Model**. In this experiment, we follow the protocol of [41], and extract 3 different features of Res3D: res5b, pool5, and prob. We again average clip features across videos and L2-normalize each feature. For each pair of videos, we compute different distances (12 distance metrics were used in [41, 16]) which results in $3 \times 12 = 36$ dimen-sional vectors. These vectors are used to train a linear SVM to predict whether the two videos contain the same action.

| Model | 2D-Resnet | AlexNet | C3D | **Res3D** |
|---|---|---|---|---|
| Accuracy (%) | 77.2 | 67.5 | 78.3 | **78.8** |
| AUC | 85.0 | 73.8 | 86.5 | **86.6** |

Table 11. **Action similarity labeling results on ASLAN**. Res3D outperforms C3D and other methods and achieves the best accuracy on ASLAN.

Table 11 presents the results of action similarity labeling using our Res3D features compared with C3D, 2D-Resnet features, as well as AlexNet features. Our Res3D features give a small improvement over C3D, but large improvements compared with 2D baselines.

### 5.3. Action detection

**Dataset**. We use the THUMOS'14 [12] temporal action localization task, a standard benchmark for action detection in long, untrimmed videos. Each video can contain multiple action instances with 20 total action categories. For training, we use all 2,755 trimmed training videos and 1,010 untrimmed validation videos (containing 3,007 action instances). For testing, we use all 213 test videos (containing 3,358 action instances) that are not entirely background videos. The temporal action localization task is to predict action instances with action categories and start/end time. Following the conventional metrics used in [12], we evaluate mean average precision (mAP) and do not allow redundant detections. A prediction is only correct if its category prediction is correct and its temporal overlap Intersection-over-Union (IoU) with the ground truth is larger than a threshold used during evaluation.

**Models**. Shou *et al.* [32] proposed a Segment-based 3D CNN framework (S-CNN) for action localization, and showed superior performance over other methods when using C3D (denoted as **C3D + S-CNN**). In this experiment we replace C3D with Res3D (denoted as **Res3D + S-CNN**) to determine whether a better spatiotemporal feature can improve performance. All 3 networks (proposal, classification, localization) of S-CNN are fine-tuned in an end-to-end manner following the protocol of [32]: the learning rate is set to 0.001 for all layers except 0.01 for the last layer; the learning rate is divided by 2 after every 4 epochs; and training is done at 16 epochs. All other settings are the same as [32]. As shown in Table 12, **Res3D + S-CNN** outperforms its direct comparable baseline, C3D+S-CNN, on THUMOS'14 (*e.g.* 3.5% mAP gain over **C3D + S-CNN** when IoU threshold is 0.5). Some baselines in [13, 43, 25, 3] are SVM classifiers trained on a pool of features that do not specifically address the detection problem. Others are based on iDT with Fisher Vector [9, 27] or RNN/LSTM [47, 48]. Unlike 3D ConvNets (C3D and Res3D), those methods cannot explicitly model appearance and motion information simultaneously. We note that there are some concurrent work with us, e.g. R-C3D [46] and CDC [31] achiving better performance on THUMOS'14. These models use C3D as their base network, we hope a similar gain can be achieved when replacing C3D by Res3D on these systems.

| IoU threshold | 0.3 | 0.4 | 0.5 | 0.6 | 0.7 |
|---|---|---|---|---|---|
| Karaman *et al*. [13] | 0.5 | 0.3 | 0.2 | 0.2 | 0.1 |
| Wang *et al*. [43] | 14.6 | 12.1 | 8.5 | 4.7 | 1.5 |
| Heilbron *et al*. [9] | - | - | 13.5 | - | - |
| Escorcia *et al*. [3] | - | - | 13.9 | - | - |
| Oneata *et al*. [25] | 28.8 | 21.8 | 15.0 | 8.5 | 3.2 |
| Richard and Gall [27] | 30.0 | 23.2 | 15.2 | - | - |
| Yeung *et al*. [47] | 36.0 | 26.4 | 17.1 | - | - |
| Yuan *et al*. [48] | 33.6 | 26.1 | 18.8 | - | - |
| Shou *et al*. [31] | 40.1 | 29.4 | 23.3 | 13.1 | 7.9 |
| Xu *et al*. [46] | **44.8** | **35.6** | **28.9** | - | - |
| C3D + S-CNN [32] | 36.3 | 28.7 | 19.0 | 10.3 | 5.3 |
| **Res3D + S-CNN** | 40.6 | 32.6 | 22.5 | 12.3 | 6.4 |

Table 12. Temporal action localization mAP on THUMOS'14 with the overlap IoU threshold used in evaluation varied from 0.3 to 0.7. - indicates that results are unavailable in the corresponding papers.

## 6. Discussion

We have presented an empirical architecture search for video classification on UCF101. We showed that our observations are useful for spatiotemporal feature learning on the large-scale Sports-1M dataset. Our proposed model, Res3D, achieves the best performance on Sports-1M, when compared with models applied on a single crop and without long-term modeling. Our Res3D outperforms C3D by a good margin across **5** different benchmarks: Sports-1M, UCF101, HMDB51, ASLAN, and THUMOS14. In addition, Res3D is **2** times faster in run-time, **2** times smaller in model size, and more compact than C3D, e.g. outperforming C3D by 5% on low dimensions on UCF101.

Although the current video benchmarks and machine capacity (e.g. GPU memory) are not ideal for ConvNet architecture search, we showed that under careful experimental settings architecture search is still valuable. This paper also provides empirical evidence, consistent with the finding in [41], that 3D convolutions are more suitable for spatiotemporal feature learning than 2D convolutions. Various video understanding applications such as action recognition, action similarity labeling, and action detection were significantly improved by using the Res3D architecture. We hope that other tasks can also benefit from Res3D, as happened with C3D. For the sake of reproducible research, the source code and pre-trained models are available at <http://github.com/facebook/C3D>.

**Acknowledgment**: we would like to thank Kaiming He and colleagues at Facebook AI Research and Applied Machine Learning for valuable feedback.

### Appendix A: Architechture Details

Table 13 and Table 14 provide further details of the architectures used in our architecture search with mixed-convolutions and varying network depth.

### Appendix B: Filter Visualizations

We include a GIF animation of our Res3D conv1 filters (<http://goo.gl/uES8Ma>). The GIF animation shows that most 3D filters can capture not only different motion patterns, but also the appearance information while 2D filters can only detect appearance information. We observe that most of the 2D filters can be matched to a temporal slice of a 3D filter. We include also the image filters of the 2D-Resnet18 baseline (2d_conv1.png) and our Res3D conv1 filters visualized in images (3d_conv1.png).

### Appendix C: Architecture Search Validation on HMDB51

**Settings**. We conduct additional experiments to verify our architecture search on HMDB51 [18]. In this section we use the same architectures as in our architecture search experiments on UCF101. Because HMDB51 is about 2 times smaller than UCF101, we adjust the training scheme accordingly. More specific, we keep the initial learning rate the same as in UCF101 which is $0.01$, but divided by $10$ at every 10K iterations (instead of 20K), and training is stopped at $45$K (instead of 90K). We use train and test split 1. As HMDB51 is smaller than UCF101, more overfitting is expected. The absolute accuracy is not important, but we would like verify if the relative ranking between architectures remains consistent with the search in UCF101. If so, we can be more confident about our empirical observations.

We re-run most of architecture search experiments as done on UCF101, except for the experiment with varying sampling rates. The main reason is that HMDB51 has many short videos, when applying on higher sampling rate (equivalent to longer input clips), the number of training and testing examples are significantly dropped. More specific, HMDB51 will loose about 20%, 45%, and 82% of its examples when input clip length is increased to 32, 64, and 128 frames (sampling rate of 8, 16, and 32), respectively.

**Results**. Table 15 presents HMDB51 accuracy of SR18 with different input resolutions. We observe that the results are consistent with our experiments on UCF101. It shows that, training on a higher resolution and with a bigger receptive field is even harder on a smaller benchmark.

Table 16 reports the accuracy of different mixed-convolution architectures on HMDB51 along with UCF101 results. We observe that all of the mixed-convolution architectures perform consistently worse than 3D, while C2.5D is comparable to 3D (SR18).

Table 17 presents the HMDB51 accuracy of different architectures with varying the network depth along with the UCF101 results. We found that the relative ranking between architectures are consistent on both datasets. This confirms that, for 3D Resnets, depth of 18 is a good trade-off for video classification.

Figure 5 plots the accuracy versus computation cost (FLOPs) for different architectures on both UCF101 and HMDB51. Different architectures are visualized by different color dots. We observe that, although the performance gaps are different, the distribtion of the color dots are consistent for UCF101 and HMDB51. This fact further confirms that under a careful design, architecture search on small benchmarks is still relevant.

| layer name | output size | MC1 | MC2 | MC3 | MC4 | MC5 |
|---|---|---|---|---|---|---|
| conv1 | 4×56×56 | $\underline{1}$ × 7 × 7, 109 | **3** × 7 × 7, 109 | **3** × 7 × 7, 108 | **3** × 7 × 7, 103 | **3** × 7 × 7, 90 |
| conv2_x | 4×56×56 | $\begin{bmatrix}\underline{1}\times3\times3, 109\\ \underline{1}\times3\times3, 109\end{bmatrix}\times2$ | $\begin{bmatrix}\underline{1}\times3\times3, 109\\ \underline{1}\times3\times3, 109\end{bmatrix}\times2$ | $\begin{bmatrix}\mathbf{3}\times3\times3, 108\\ \mathbf{3}\times3\times3, 108\end{bmatrix}\times2$ | $\begin{bmatrix}\mathbf{3}\times3\times3, 103\\ \mathbf{3}\times3\times3, 103\end{bmatrix}\times2$ | $\begin{bmatrix}\mathbf{3}\times3\times3, 90\\ \mathbf{3}\times3\times3, 90\end{bmatrix}\times2$ |
| conv3_x | 2×28×28 | $\begin{bmatrix}\underline{1}\times3\times3, 218\\ \underline{1}\times3\times3, 218\end{bmatrix}\times2$ | $\begin{bmatrix}\underline{1}\times3\times3, 218\\ \underline{1}\times3\times3, 218\end{bmatrix}\times2$ | $\begin{bmatrix}\underline{1}\times3\times3, 216\\ \underline{1}\times3\times3, 216\end{bmatrix}\times2$ | $\begin{bmatrix}\mathbf{3}\times3\times3, 206\\ \mathbf{3}\times3\times3, 206\end{bmatrix}\times2$ | $\begin{bmatrix}\mathbf{3}\times3\times3, 180\\ \mathbf{3}\times3\times3, 180\end{bmatrix}\times2$ |
| conv4_x | 1×14×14 | $\begin{bmatrix}\underline{1}\times3\times3, 436\\ \underline{1}\times3\times3, 436\end{bmatrix}\times2$ | $\begin{bmatrix}\underline{1}\times3\times3, 436\\ \underline{1}\times3\times3, 436\end{bmatrix}\times2$ | $\begin{bmatrix}\underline{1}\times3\times3, 432\\ \underline{1}\times3\times3, 432\end{bmatrix}\times2$ | $\begin{bmatrix}\underline{1}\times3\times3, 412\\ \underline{1}\times3\times3, 412\end{bmatrix}\times2$ | $\begin{bmatrix}\mathbf{3}\times3\times3, 360\\ \mathbf{3}\times3\times3, 360\end{bmatrix}\times2$ |
| conv5_x | 1×7×7 | $\begin{bmatrix}\underline{1}\times3\times3, 872\\ \underline{1}\times3\times3, 872\end{bmatrix}\times2$ | $\begin{bmatrix}\underline{1}\times3\times3, 872\\ \underline{1}\times3\times3, 872\end{bmatrix}\times2$ | $\begin{bmatrix}\underline{1}\times3\times3, 864\\ \underline{1}\times3\times3, 864\end{bmatrix}\times2$ | $\begin{bmatrix}\underline{1}\times3\times3, 824\\ \underline{1}\times3\times3, 824\end{bmatrix}\times2$ | $\begin{bmatrix}\underline{1}\times3\times3, 720\\ \underline{1}\times3\times3, 720\end{bmatrix}\times2$ |
| pool5 | 1×1×1 | average pool, 101-d fc, softmax | | | | |
| FLOPs ($\times 10^9$) | | 10.2 | 9.7 | 20.8 | 22.6 | 18.7 |
| # of parameters ($\times 10^6$) | | 32.5 | 32.5 | 32.8 | 32.4 | 33.0 |

Table 13. **Resnet architectures with different mixtures of 3D and 2D convolutions**. Building blocks are shown in brackets, with the numbers of blocks stacked. Downsampling is performed by conv1 and conv5_1 with a stride of $1 \times 2 \times 2$, and conv3_1 and conv4_1 with a stride of $2 \times 2 \times 2$. For better contrasting the difference, 3D convolutions and 2D convolutions are presented in bold red and underlined blue text, repsectively.

| layer name | output size | D10 | D16 | D18* | D26 | D34 |
|---|---|---|---|---|---|---|
| conv1 | 4×56×56 | 3 × 7 × 7, $\underline{98}$ | 3 × 7 × 7, $\underline{85}$ | 3 × 7 × 7, $\underline{64}$ | 3 × 7 × 7, $\underline{50}$ | 3 × 7 × 7, $\underline{46}$ |
| conv2_x | 4×56×56 | $\begin{bmatrix}3\times3\times3, \underline{98}\\ 3\times3\times3, \underline{98}\end{bmatrix}\times\mathbf{1}$ | $\begin{bmatrix}3\times3\times3, \underline{85}\\ 3\times3\times3, \underline{85}\end{bmatrix}\times\mathbf{2}$ | $\begin{bmatrix}3\times3\times3, \underline{64}\\ 3\times3\times3, \underline{64}\end{bmatrix}\times\mathbf{2}$ | $\begin{bmatrix}3\times3\times3, \underline{50}\\ 3\times3\times3, \underline{50}\end{bmatrix}\times\mathbf{2}$ | $\begin{bmatrix}3\times3\times3, \underline{46}\\ 3\times3\times3, \underline{46}\end{bmatrix}\times\mathbf{3}$ |
| conv3_x | 2×28×28 | $\begin{bmatrix}3\times3\times3, \underline{196}\\ 3\times3\times3, \underline{196}\end{bmatrix}\times\mathbf{1}$ | $\begin{bmatrix}3\times3\times3, \underline{170}\\ 3\times3\times3, \underline{170}\end{bmatrix}\times\mathbf{2}$ | $\begin{bmatrix}3\times3\times3, \underline{128}\\ 3\times3\times3, \underline{128}\end{bmatrix}\times\mathbf{2}$ | $\begin{bmatrix}3\times3\times3, \underline{100}\\ 3\times3\times3, \underline{100}\end{bmatrix}\times\mathbf{3}$ | $\begin{bmatrix}3\times3\times3, \underline{92}\\ 3\times3\times3, \underline{92}\end{bmatrix}\times\mathbf{4}$ |
| conv4_x | 1×14×14 | $\begin{bmatrix}3\times3\times3, \underline{392}\\ 3\times3\times3, \underline{392}\end{bmatrix}\times\mathbf{1}$ | $\begin{bmatrix}3\times3\times3, \underline{340}\\ 3\times3\times3, \underline{340}\end{bmatrix}\times\mathbf{2}$ | $\begin{bmatrix}3\times3\times3, \underline{256}\\ 3\times3\times3, \underline{256}\end{bmatrix}\times\mathbf{2}$ | $\begin{bmatrix}3\times3\times3, \underline{200}\\ 3\times3\times3, \underline{200}\end{bmatrix}\times\mathbf{4}$ | $\begin{bmatrix}3\times3\times3, \underline{184}\\ 3\times3\times3, \underline{184}\end{bmatrix}\times\mathbf{6}$ |
| conv5_x | 1×7×7 | $\begin{bmatrix}3\times3\times3, \underline{784}\\ 3\times3\times3, \underline{784}\end{bmatrix}\times\mathbf{1}$ | $\begin{bmatrix}3\times3\times3, \underline{680}\\ 3\times3\times3, \underline{680}\end{bmatrix}\times\mathbf{1}$ | $\begin{bmatrix}3\times3\times3, \underline{512}\\ 3\times3\times3, \underline{512}\end{bmatrix}\times\mathbf{2}$ | $\begin{bmatrix}3\times3\times3, \underline{400}\\ 3\times3\times3, \underline{400}\end{bmatrix}\times\mathbf{3}$ | $\begin{bmatrix}3\times3\times3, \underline{368}\\ 3\times3\times3, \underline{368}\end{bmatrix}\times\mathbf{3}$ |
| pool5 | 1×1×1 | average pool, 101-d fc, softmax | | | | |
| FLOPs ($\times 10^9$) | | 10.8 | 16.7 | 10.3 | 8.4 | 10.0 |
| # of parameters ($\times 10^6$) | | 33.7 | 33.6 | 33.2 | 33.8 | 32.8 |

Table 14. **3D-Resnet architectures with different number of layers**. Building blocks are shown in brackets, with the numbers of blocks stacked. Downsampling is performed by conv1 and conv5_1 with a stride of $1 \times 2 \times 2$, and conv3_1 and conv4_1 with a stride of $2 \times 2 \times 2$. *D18 is equivalent to SR18. For better contrasting the difference, number of filters and number of blocks are presented in bold red and underlined blue text, repsectively.

| Input resolution (crop size) | 64 (56) | 128 (112) | 256 (224) |
|---|---|---|---|
| UCF101 | 37.6 | 46.1 | 44.0 |
| HMDB51 | 18.3 | 21.5 | 17.9 |

Table 15. **Accuracy on UCF101 and HMDB, varying input resolution**. The experimental result on HMDB51 is consistent with the observation on UCF101. This confirms resolution 128 is a good choice for video classification.

| Net | MC1 | MC2 | MC3 | MC4 | MC5 | 2.5D | 3D |
|---|---|---|---|---|---|---|---|
| UCF101 | 43.3 | 44.5 | 44.8 | 44.3 | 45.1 | 44.6 | 46.1 |
| HMDB51 | 19.7 | 19.4 | 19.9 | 20.4 | 20.9 | 21.4 | 21.5 |

Table 16. **Accuracy of different mixed-convolution architectures accuracy on UCF101 and HMDB51**. We further verify the observations on HMDB51 and confirm that most of the observations are also hold for HMDB51.

| Net | D10 | D16 | D18 | D26 | D34 |
|---|---|---|---|---|---|
| UCF101 | 45.3 | 46.8 | 46.1 | 46.3 | 45.8 |
| HMDB51 | 21.3 | 22.8 | 21.5 | 22.0 | 20.1 |

Table 17. **Accuracy of different depth architectures on UCF101 and HMDB51**. On HMDB51, D34 performs slightly worse than the others, D16 and D26 are also among the best, which is consistent with UCF101. D18 performs slightly worse than D16 and D26 but at lower computation complexity and less memory requirement.

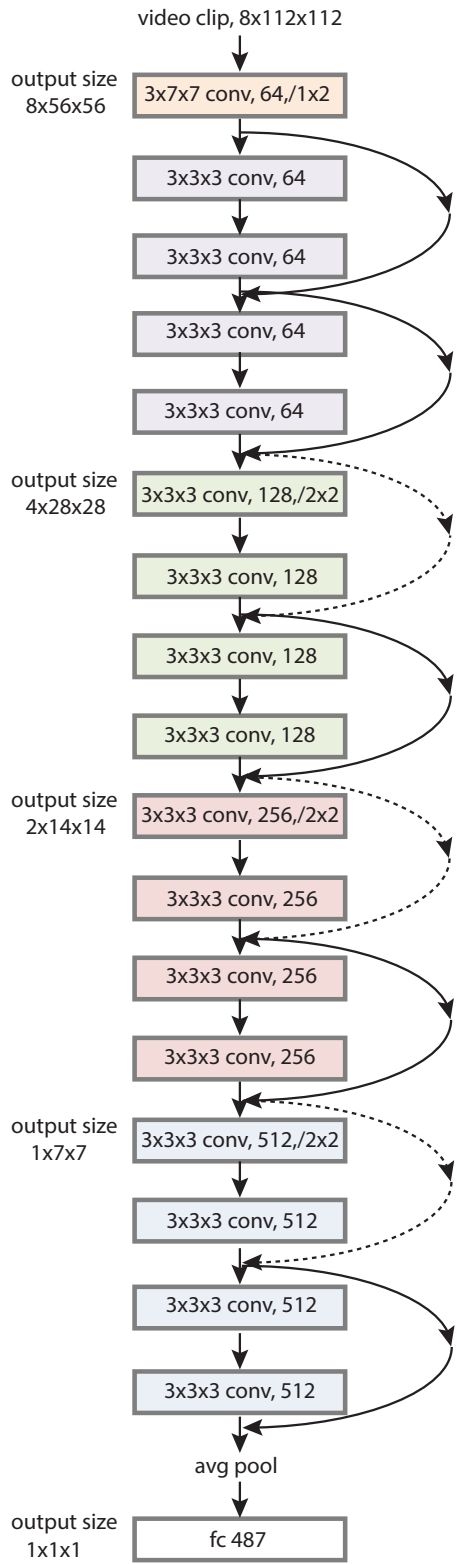

Figure 4. **Res3D architecture**. Downsampling strides are denoted as $t \times s$ where $t$ and $s$ are temporal and spatial stride, respectively. Dotted lines are residual connections with downsampling.

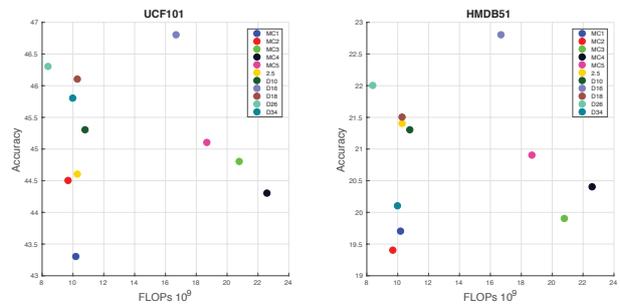

Figure 5. **Architecture search on UCF101 vs. HMDB51**. We plot the accuracy versus computation (FLOPs) for different architectures on UCF101 and HMDB51. The left plot shows results on UCF101 while the right plot shows the results on HMBD51. Different colors are used for different architectures. The distribution of the color dots is consistent for UCF101 and HMDB51.